\documentclass[letterpaper, 10 pt, conference]{ieeeconf}
\IEEEoverridecommandlockouts 
\usepackage[english]{babel}
\usepackage{textcomp}
\usepackage{amsmath}
\usepackage{amssymb}
\usepackage{booktabs}
\usepackage{caption}
\usepackage{subfigure}
\usepackage{multirow}
\usepackage{amsfonts}
\usepackage{graphicx}
\usepackage{breakurl}
\usepackage{enumerate}
\usepackage{tabularx}
\usepackage{algorithmic}
\usepackage{algorithm}
\usepackage{bm}
\usepackage{subcaption}
\usepackage{enumerate}
\usepackage{adjustbox}
\usepackage{xcolor}
\usepackage{siunitx}
\usepackage{booktabs}
\usepackage{mdframed}
\usepackage{adjustbox}
\usepackage{color}
\usepackage{xurl}
\usepackage{makecell}
\usepackage{cite}
\makeatletter
\let\NAT@parse\undefined
\makeatother
\usepackage{hyperref}
\usepackage{relsize}
\usepackage{float}
\usepackage{pifont}

\title{\LARGE \bf
HUNT: High-Speed UAV Navigation and Tracking in\\Unstructured Environments via Instantaneous Relative Frames
}

\author{Alessandro Saviolo$^1$, Jeffrey Mao$^1$, and Giuseppe Loianno$^2$%
\thanks{
$^1$The authors are with New York University, NY 10012, USA. {\tt\footnotesize email: \{as16054, jm7752\}@nyu.edu}.}
\thanks{
$^2$The author is with University of California Berkeley, Department of Electrical Engineering and Computer Sciences, Berkeley, CA 94720, USA. {\tt\footnotesize email: loiannog@eecs.berkeley.edu}.}
\thanks{This work was supported by the NSF CAREER Award 2546659, the DARPA YFA Grant D22AP00156-00, and the DEVCOM ARL grant SARA W911NF-24-2-0057.}
}

\begin{document}

\thispagestyle{empty}
\pagestyle{empty}

\maketitle

\begin{abstract}
Search and rescue operations require unmanned aerial vehicles to both traverse unknown unstructured environments at high speed and track targets once detected. Achieving both capabilities under degraded sensing and without global localization remains an open challenge. Recent works on relative navigation have shown robust tracking by anchoring planning and control to a visible detected object, but cannot address navigation when no target is in the field of view.  
We present \textbf{HUNT} (High-speed UAV Navigation and Tracking), a real-time framework that unifies traversal, acquisition, and tracking within a single relative formulation. HUNT defines navigation objectives directly from onboard instantaneous observables such as attitude, altitude, and velocity, enabling reactive high-speed flight during search. Once a target is detected, the same perception–control pipeline transitions seamlessly to tracking. Outdoor experiments in dense forests, container compounds, and search-and-rescue operations with vehicles and mannequins demonstrate robust autonomy where global methods fail.
\textbf{Video}: \url{https://youtu.be/YsSflqPPHhs}
\end{abstract}

\section{Introduction}
Unmanned Aerial Vehicles (UAVs), especially quadrotors, have become essential tools in Search-and-Rescue (SAR) operations, where their maneuverability and speed enable rapid deployment in unstructured, GPS-denied environments. These missions require two critical capabilities: the ability to \textit{safely traverse unknown terrain at high speed} to search for victims or objects of interest, and the ability to \textit{reliably track and follow detected targets} once they are identified. Achieving both in degraded sensing and unknown environments remains a fundamental challenge in robotics.

Traditional autonomy relies on global navigation anchored by GPS, Visual-Inertial Odometry (VIO), or Simultaneous Localization and Mapping (SLAM)~\cite{mao2025time}. These methods degrade in exactly the conditions most critical to SAR: dense forests, collapsed buildings, or urban canyons where GPS is degraded and visual features are sparse, dynamic, or occluded. Their dependence on persistent landmarks and loop closures makes global pose estimation fragile, resulting in drift and unsafe behavior at high speed.

An alternative paradigm, recently advanced in prior works \cite{saviolo2025nova}, is to abandon global consistency and instead define navigation directly in a target-relative reference frame. By continuously re-anchoring perception, planning, and control to the instantaneous detection of a moving object, UAVs can robustly pursue targets even in unstructured or degraded environments. This \textit{instantaneous relative navigation} formulation has demonstrated strong performance for the tracking phase of SAR missions. However, it presumes that a target is visible from the outset and remains continuously observable—an assumption that rarely holds in practice. What is missing is a capability for safe, high-speed traversal in unknown terrain \textit{without any target directly in view}, and without reverting to the fragile assumptions of traditional GPS, VIO, or SLAM.

\begin{figure}[t]
    \centering
    \includegraphics[width=\linewidth]{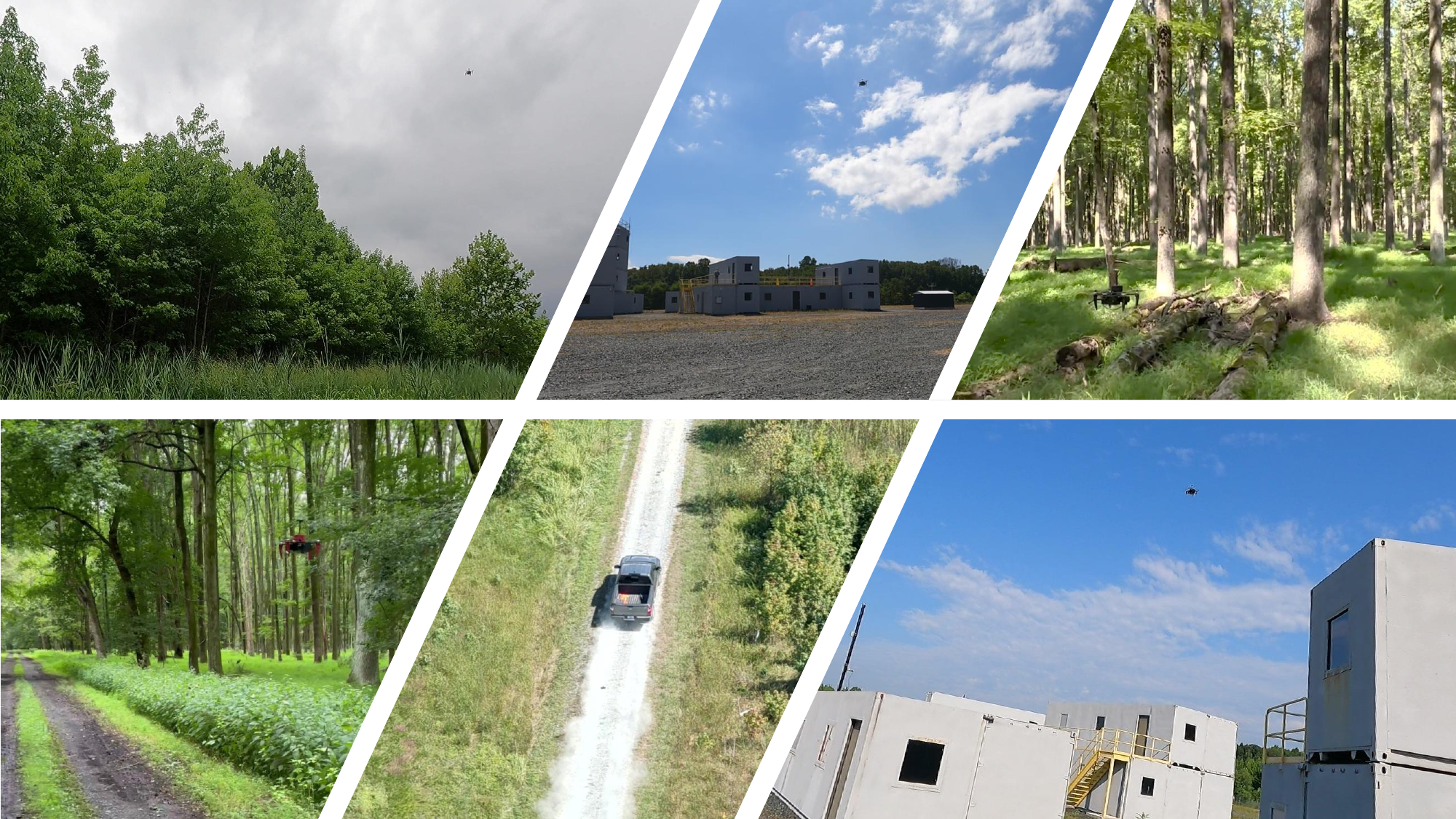}
    \caption{\textbf{HUNT enables reactive high-speed autonomy for search-and-rescue operations.} 
    The proposed framework is validated across diverse outdoor missions, from urban compounds to dense forests, demonstrating seamless integration of fast \textit{traversal}, target \textit{acquisition}, and sustained \textit{tracking}.}
    \label{fig:initial_figure}
    \vspace{-1em}
\end{figure}

This paper introduces \textbf{HUNT}, a reactive autonomy framework that unifies high-speed traversal, acquisition, and tracking under a single instantaneous relative formulation. At its core is a novel \textit{loitering mode}, which defines navigation objectives solely from directly observable quantities—attitude, altitude, and instantaneous velocity—allowing the UAV to traverse unknown cluttered environments safely even in the absence of a target. When a target is detected, the same perception–control pipeline seamlessly re-anchors the reference frame to the target, turning tracking into a special case of loitering without requiring global state.  

Safety is enforced throughout by embedding high-order Control Barrier Functions (CBFs) directly into a Nonlinear Model Predictive Controller (NMPC), guaranteeing dynamically feasible, collision-free trajectories in real time. Robust transitions between modes are enabled by a confidence-based switching mechanism that prevents oscillations and ensures stability as detections appear and disappear.  

We validate HUNT in extensive outdoor SAR-like missions spanning urban compounds, semi-structured layouts, and dense forests. Experiments include loitering flights over city blocks and forest canopy, dense-clutter traversal below canopy, and full missions where the UAV searches, acquires, and pursues static and dynamic targets such as vehicles and mannequins. Across all settings, HUNT achieves consistent high-speed performance where global methods fail.  

\textit{To the best of our knowledge, HUNT is the first framework to integrate traversal, acquisition, and tracking in cluttered, GPS-denied environments under a unified instantaneous relative formulation, eliminating reliance on global pose.}

\section{Related Works}
Research on GPS-denied navigation can be broadly organized by the choice of reference frame. Early approaches adopted a \textit{world-centric} formulation, estimating and controlling motion relative to a fixed global frame. VIO and Visual-Inertial SLAM (VI-SLAM) exemplify this class, fusing relative measurements over time to reconstruct trajectories~\cite{ORBSLAM3_TRO}. While accurate in structured environments with abundant features and loop closures, they degrade rapidly at high speed or under feature sparsity. LiDAR integration has improved robustness~\cite{lio2}, but such systems remain heavy, energy-inefficient, and fundamentally incompatible with agile flight. They are also susceptible to jamming~\cite{lidar_jam} and interference in multi-agent scenarios~\cite{lidar_crosstalk}. Landmark-based localization~\cite{jiuhong_geolocalization} can reduce computation and support agile flight~\cite{li2020visual}, yet presumes structured known environments unsuitable for SAR.

In contrast, \textit{object-centric} formulations discard global consistency and express navigation relative to detected objects. A canonical example is Proportional Navigation~\cite{pn_fudamentals}, used since the 1950s for missile interception at extreme speeds. Variants have been deployed on quadrotors for agile pursuit~\cite{saviolo2024unifying} and adapted to camera-based control through image-based visual servoing for precision landing~\cite{thomas2015visual}. These methods relax environmental assumptions and support fast maneuvers, but are fragile to target loss when objects leave the field of view and cannot directly handle constraints.

Several efforts have sought to mitigate the loss of target visibility. Visibility-constrained methods maintain the target in the field of view~\cite{mao_robust} but require prior detection and cannot ensure observability without the object's trajectory knowledge. Cooperative approaches leverage state broadcasts from tracked agents~\cite{ahmad2022pacnav}, but applicability is limited to collaborative scenarios. Maintaining a library of landmarks for relative navigation has been explored through online factor graphs~\cite{relative_fixed_wing} or predefined libraries~\cite{maggio2023vision}, though these reintroduce world-centric challenges or demand prior maps. Open-loop fallback maneuvers have been demonstrated in structured contexts like competitive drone racing~\cite{scarciglia2025map}, and hybrid systems that switch between world- and object-centric formulations~\cite{AIVIO} improve overall system robustness but still suffer from the fragility of global pose estimation.

For handling constraints, optimization-based extensions have incorporated environmental constraints into relative navigation. Cooperative optimization has been demonstrated at low speeds with continuous state sharing~\cite{zhang2023coni}, while LiDAR-augmented schemes~\cite{zhang2025global} expand awareness but assume static environments and modest velocities ($<1.5~\si{m/s}$). Learning- and NMPC-based frameworks~\cite{saviolo2025nova} address tracking in unstructured environments at higher speeds, yet all existing approaches remain fundamentally limited by their reliance on continuous target visibility.

\begin{figure}[t]
    \centering
    \includegraphics[width=\linewidth, trim=0 250 435 0, clip]{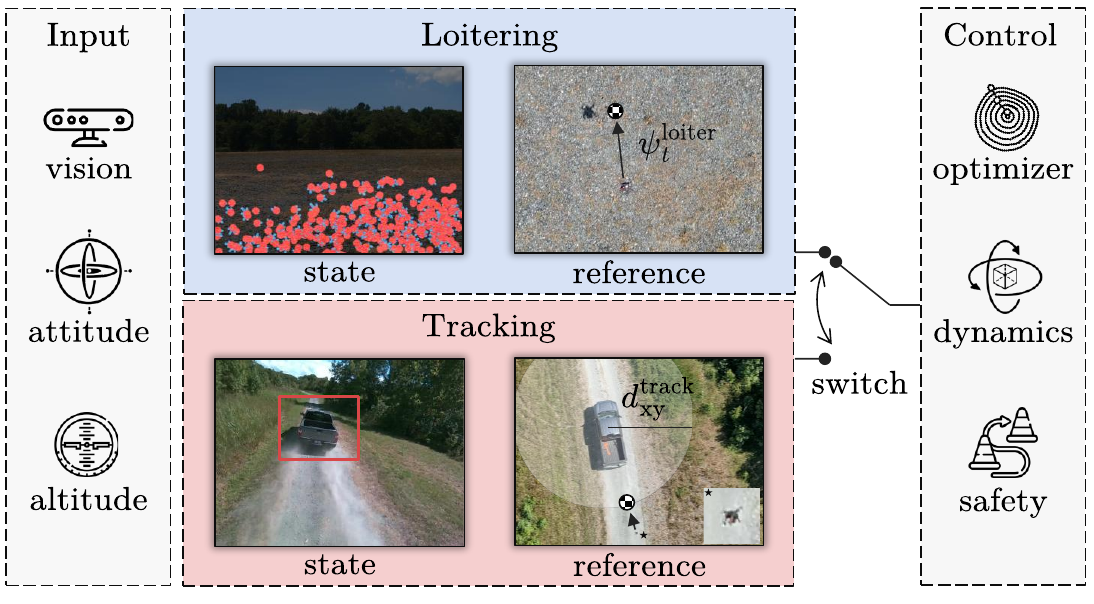}
    \caption{\textbf{HUNT framework overview.} 
    Inputs from onboard vision, attitude, and altitude sensors are fused to reconstruct directly observable states in the instantaneous relative frame~$\mathcal{R}_t$. 
    In \textit{loitering}, $\mathcal{R}_t$ is anchored to gravity and the initial heading, and the reference trajectory regulates altitude, heading ($\psi_{t}^{\text{loiter}}$), and forward velocity. 
    In \textit{tracking}, $\mathcal{R}_t$ is re-anchored on the detected target, and the reference trajectory enforces a safety distance ($d_{xy}^{\text{track}}$) while continuously orienting toward it. 
    A unified NMPC optimizes future trajectories under dynamics, actuation, and obstacle-avoidance constraints, with seamless switching between modes.}
    \label{fig:method}
    \vspace{-1em}
\end{figure}

\section{Methodology}
HUNT formulates perception, estimation, and control entirely within an \textit{instantaneous relative frame}~$\mathcal{R}_t$. This frame is reconstructed at every control cycle $t$ from directly measurable quantities—attitude, altitude, instantaneous velocity, and relative target position when available—thereby eliminating dependence on unobservable states such as horizontal translation or absolute yaw. By anchoring objectives and constraints in~$\mathcal{R}_t$, the system preserves the essential geometric structure required for constraint-aware control while avoiding the drift and fragility of global pose estimation.

Two operating regimes emerge naturally (Figure~\ref{fig:method}). In \textit{loitering}, the UAV regulates its motion in an environment-anchored frame to traverse cluttered terrain while searching for a target. In \textit{tracking}, the frame origin is re-anchored at each cycle to the detected target, yielding consistent pursuit. Both regimes are implemented within a single perception–control pipeline. The following subsections detail the construction of~$\mathcal{R}_t$, the perception modules that supply instantaneous observables, and the control architecture that enforces dynamic feasibility and safety.

\subsection{Loitering Mode}
Loitering corresponds to the navigation regime where no target is detected with high confidence. In this setting, the UAV must sustain safe high-speed flight through cluttered terrain while actively searching, without relying on global localization or precomputed maps. The instantaneous frame $\mathcal{R}_t$ is anchored to gravity and the initial heading and then held fixed over time. Control in this regime depends only on states that are directly observable onboard and consistently mapped from the body frame~$\mathcal{B}$ into $\mathcal{R}_t$, ensuring robustness even as unobservable global variables drift.  

\subsubsection{State Estimation}
State estimation is performed with an Unscented Kalman Filter (UKF) that maintains the state
\begin{equation}
    \mathbf{x}_t =
    \begin{bmatrix}
        \mathbf{p}_t^{\mathcal{R}_t \mathcal{B}} \\
        \mathbf{v}_t^{\mathcal{R}_t \mathcal{B}} \\
        \mathbf{q}_t^{\mathcal{R}_t \mathcal{B}} \\
        b_{p,t} \\
        \mathbf{b}_{a,t} \\
        \mathbf{b}_{g,t}
    \end{bmatrix},
\end{equation}
where $\mathbf{p}_t^{\mathcal{R}_t \mathcal{B}}$ is the relative position, $\mathbf{v}_t^{\mathcal{R}_t \mathcal{B}}$ the velocity, $\mathbf{q}_t^{\mathcal{R}_t \mathcal{B}}$ the attitude of $\mathcal{B}$ with respect to $\mathcal{R}_t$, and $b_{p,t}, \mathbf{b}_{a,t}, \mathbf{b}_{g,t}$ the barometer, accelerometer, and gyroscope biases.  

The state is propagated using IMU readings
\begin{equation}
    \mathbf{z}_t^{\mathrm{process}} =
    \begin{bmatrix}
        \mathbf{a}_t^{\mathcal{B}} \\
        \boldsymbol{\omega}_t^{\mathcal{B}}
    \end{bmatrix},
\end{equation}
through inertial kinematics. Position $\mathbf{p}_t^{\mathcal{R}_t \mathcal{B}}$ is obtained by integrating velocity, $\mathbf{v}_t^{\mathcal{R}_t \mathcal{B}}$ by integrating accelerations $\mathbf{a}_t^{\mathcal{B}}$, and $\mathbf{q}_t^{\mathcal{R}_t \mathcal{B}}$ by integrating angular rates $\boldsymbol{\omega}_t^{\mathcal{B}}$.

The state is updated with the measurement vector
\begin{equation}
    \mathbf{z}_t^{\mathrm{update}} =
    \begin{bmatrix}
        p_{z,t}^{\mathcal{R}_t \mathcal{B}} \\
        \mathbf{v}_t^{\mathcal{R}_t \mathcal{B}}
    \end{bmatrix},
\end{equation}
whose first component corresponds to the barometric altitude and second to the velocity reconstructed from stereo–inertial fusion. Specifically, velocity is obtained by tracking optical flow features between two sequential frames~\cite{alfarano2024estimating}, rejecting outliers via essential-matrix RANSAC, and recovering metric scale from disparity. The resulting velocity is corrected for the camera lever arm, rotated into $\mathcal{B}$, and mapped into $\mathcal{R}_t$.

This choice of state follows the traditional formulation for UAV dynamics, allowing us to leverage decades of results in modeling, estimation, and control~\cite{saviolo2023learning}. At the same time, it unavoidably introduces drift in the horizontal position components $p_{x,t}, p_{y,t}$, since they are not directly observable. In HUNT, this drift does not compromise navigation: reference generation (Subsection~\ref{subsubsec:loiter_reference_gen}) always defines goals relative to the current state estimate, ensuring that control objectives remain well-posed even as horizontal position diverges.

\subsubsection{Reference Generation} \label{subsubsec:loiter_reference_gen}
Given the estimated state $\mathbf{x}_t$, loitering generates references in terms of altitude $p_z^{\mathrm{loiter}}$, forward speed $v^{\mathrm{loiter}}$, and heading $\psi_t^{\mathrm{loiter}}$. The altitude reference regulates vertical motion, the forward speed sets the cruise velocity, and the heading defines the direction of travel.

The heading $\psi_t^{\mathrm{loiter}}$ maintains the initial yaw while continuously running the object detector. Low-confidence detections bias the heading toward the projected center of the bounding box, providing a weak directional cue without triggering a mode switch. Only when detection confidence exceeds the switching threshold does the system transition to tracking, as detailed in Subsection~\ref{subsec:switching_mode}.

At each stage $i \in [0,N)$ of the NMPC prediction horizon of length $N$ and discretization step $\Delta t$, the reference trajectory $\{\bar{\mathbf{x}}_{t+i}\}$ is rolled out by advancing at constant speed $v^{\mathrm{loiter}}$ along $\psi_t^{\mathrm{loiter}}$ while holding altitude at $p_z^{\mathrm{loiter}}$. Formally,
\begin{align}
    \bar{\mathbf{p}}_{t+i} &=
    \begin{bmatrix}
        p_{x,t} + (i\, v^{\mathrm{loiter}}\, \Delta t) \cos\psi_t^{\mathrm{loiter}} \\
        p_{y,t} + (i\, v^{\mathrm{loiter}}\, \Delta t) \sin\psi_t^{\mathrm{loiter}} \\
        p_z^{\mathrm{loiter}}
    \end{bmatrix}, \\
    \bar{\mathbf{q}}_{t+i} &=
    \begin{bmatrix}
        \cos\!\left(\tfrac{\psi_t^{\mathrm{loiter}}}{2}\right) \\
        0 \\
        0 \\
        \sin\!\left(\tfrac{\psi_t^{\mathrm{loiter}}}{2}\right)
    \end{bmatrix},
\end{align}
corresponding to flight aligned with $\psi_t^{\mathrm{loiter}}$. Because the horizontal position $p_{x,t}, p_{y,t}$ is included in the reference state, even if it drifts, the reference is defined relative to the current estimate, ensuring drift does not alter the commanded goal.

\subsection{Tracking Mode}
Tracking corresponds to the navigation regime where a target is detected with high confidence. In this setting, the UAV re-anchors the instantaneous frame $\mathcal{R}_t$ at every cycle, placing its origin at the target position. This construction keeps the target stationary in $\mathcal{R}_t$ regardless of global drift, ensuring that objectives are consistently defined in target-relative coordinates. Control in this mode aims to maintain a safety distance from the target, providing horizontal and vertical offsets while keeping the vehicle oriented toward it.

\subsubsection{State Estimation}
State estimation in tracking is performed with a UKF maintaining the same state vector as in loitering. The process model is driven by IMU angular rates
\begin{equation}
    \mathbf{z}_t^{\mathrm{process}} =
    \begin{bmatrix}
        \boldsymbol{\omega}_t^{\mathcal{B}}
    \end{bmatrix},
\end{equation}
since linear accelerations cannot be used to avoid placing any assumption about the knowledge of the target’s motion. Attitude $\mathbf{q}_t^{\mathcal{R}_t \mathcal{B}}$ is therefore propagated from gyroscope readings, while position evolves by integrating velocity $\mathbf{v}_t^{\mathcal{R}_t \mathcal{B}}$.

The state is updated using target-relative displacement obtained from visual detection. A lightweight onboard detector with adaptive zoom runs on the RGB stream; crops tighten at high confidence and expand when uncertain to improve reacquisition (as detailed in prior works on target tracking~\cite{saviolo2025nova}). Fused with stereo disparity, the detector output provides the target’s 3D displacement in the camera frame, which is corrected for the camera lever arm, rotated into $\mathcal{B}$, and mapped into $\mathcal{R}_t$. The resulting measurement vector is
\begin{equation}
    \mathbf{z}_t^{\mathrm{update}} =
    \begin{bmatrix}
        \mathbf{p}_{t}^{\mathcal{R}_t \mathcal{B}}
    \end{bmatrix},
\end{equation}
where $\mathbf{p}_{t}^{\mathcal{R}_t \mathcal{B}}$ is the detected target position expressed in $\mathcal{R}_t$.  

With this update, the UKF re-anchors the state to the target at every cycle. State and covariance are transported consistently into the moving frame, ensuring stable estimation.

\subsubsection{Reference Generation}
Given the estimated state $\mathbf{x}_t$, tracking generates references in terms of a horizontal standoff distance $d_{\mathrm{xy}}^{\mathrm{track}}$, a vertical offset $d_{\mathrm{z}}^{\mathrm{track}}$, and a heading $\psi_t^{\mathrm{track}}$ that aligns the UAV with the target. The standoff regulates lateral separation, the vertical offset fixes relative altitude, and the heading ensures the target remains centered.  

The heading $\psi_t^{\mathrm{track}}$ is updated at every cycle from the bearing of the detected target, reorienting the body $x$-axis to point toward it. This maintains a consistent look-at behavior.  

At each stage $i \in [0,N)$ of the NMPC prediction horizon of length $N$ and discretization step $\Delta t$, the reference trajectory $\{\bar{\mathbf{x}}_{t+i}\}$ is rolled out by holding the desired displacement and orientation fixed relative to the target. Formally,
\begin{align}
    \bar{\mathbf{p}}_{t+i} &=
    \begin{bmatrix}
        p_{x,t}^{\mathcal{R}_t \mathcal{B}} \\
        p_{y,t}^{\mathcal{R}_t \mathcal{B}} \\
        0
    \end{bmatrix}
    \frac{d_{\mathrm{xy}}^{\mathrm{track}}}{
        \sqrt{(p_{x,t}^{\mathcal{R}_t\mathcal{B}})^2 + (p_{y,t}^{\mathcal{R}_t\mathcal{B}})^2}
    }
    +
    \begin{bmatrix}
        0 \\ 0 \\ d_{\mathrm{z}}^{\mathrm{track}}
    \end{bmatrix}, \\
    \bar{\mathbf{q}}_{t+i} &=
    \begin{bmatrix}
        \cos\!\left(\tfrac{\psi_t^{\mathrm{track}}}{2}\right) \\
        0 \\
        0 \\
        \sin\!\left(\tfrac{\psi_t^{\mathrm{track}}}{2}\right)
    \end{bmatrix}.
\end{align}

Since references are always expressed relative to the target in $\mathcal{R}_t$, horizontal drift has no effect.

\subsection{Collision Avoidance} \label{subsection:collision_avoidance}
Collision avoidance is enforced by embedding CBF constraints directly in $\mathcal{R}_t$. At each cycle, stereo depth images are completed, projected into a point cloud~\cite{saviolo2025nova}, and the $K$ most imminent collision points are selected using a time-to-collision heuristic~\cite{saviolo2025reactive}. Each sample $k$ defines a local frame $\mathcal{O}_{k,t}$ and is first expressed in the camera frame $\mathcal{C}$ as $\mathbf{o}_{k,t}^{\mathcal{C}\mathcal{O}_{k,t}}$. Using the calibrated extrinsics, this vector is mapped into the body frame $\mathcal{B}$ and rotated into $\mathcal{R}_t$ with $\mathbf{q}_t^{\mathcal{R}_t \mathcal{B}}$, yielding $\mathbf{o}_{k,t}^{\mathcal{R}_t \mathcal{O}_{k,t}}$.

For each candidate obstacle, the CBF encodes a quadratic separation constraint of the form
\begin{equation}
    h_k(\mathbf{x}_t) =
    \big\|\mathbf{p}_t^{\mathcal{R}_t \mathcal{B}} - \mathbf{o}_{k,t}^{\mathcal{R}_t \mathcal{O}_{k,t}}\big\|^2 - r_{\mathrm{safe}}^2 ,
\end{equation}
where $\mathbf{p}_t^{\mathcal{R}_t \mathcal{B}}$ is the UAV position in $\mathcal{R}_t$ and $r_{\mathrm{safe}}$ a safety radius. The admissible control set is defined by requiring 
\begin{equation}
    \ddot{h}_k(\mathbf{x}_t, \mathbf{u}_t) + 2\lambda \dot{h}_k(\mathbf{x}_t) + \lambda^2 h_k(\mathbf{x}_t) \ge 0,
\end{equation}
for all $k$, ensuring forward invariance of the safe set and preventing collisions; $\lambda > 0$ is a tunable barrier rate, with larger values enforcing safety more aggressively and smaller values yielding smoother but less conservative behavior.

\subsection{Motion Planning and Control}
Both loitering and tracking are executed through a single unified controller: an NMPC augmented with CBF constraints. At each cycle, the NMPC solves a finite-horizon problem of length $N$, minimizing the deviation between the mode-specific predicted states $\hat{\mathbf{x}}_{t+i}$ and references $\bar{\mathbf{x}}_{t+i}$:
\begin{equation}
    \min_{\substack{\hat{\mathbf{x}}_{t}, \dots, \hat{\mathbf{x}}_{t+N} \\ \hat{\mathbf{u}}_{t}, \dots, \hat{\mathbf{u}}_{t+N-1}}}
    \sum_{j=0}^{N} 
    \left\| \hat{\mathbf{x}}_{t+j} - \bar{\mathbf{x}}_{t+j} \right\|_\mathbf{Q}^2
    + \sum_{j=0}^{N-1} 
    \left\| \hat{\mathbf{u}}_{t+j} \right\|_\mathbf{R}^2 ,
\end{equation}
where $\mathbf{Q}$, $\mathbf{R}$ are positive definite weighting matrices penalizing tracking error and control effort.  
The optimization is subject to dynamics, input bounds, and safety constraints for all predictions $j \in [0,N)$ and obstacles $k \in [0,K)$:
\begin{subequations}
\begin{align}
    &\hat{\mathbf{x}}_{t} = \mathbf{x}_{t}, \\
    &\hat{\mathbf{x}}_{t+1+j} = f(\hat{\mathbf{x}}_{t+j}, \hat{\mathbf{u}}_{t+j}), \\
    &\mathbf{x}_{\min} \leq \hat{\mathbf{x}}_{t+j} \leq \mathbf{x}_{\max}, \\
    &\mathbf{u}_{\min} \leq \hat{\mathbf{u}}_{t+j} \leq \mathbf{u}_{\max}, \\
    &\ddot{h}_k(\hat{\mathbf{x}}_{t+j}, \hat{\mathbf{u}}_{t+j}) 
    + 2 \lambda \dot{h}_k(\hat{\mathbf{x}}_{t+j}) 
    + \lambda^2 h_k(\hat{\mathbf{x}}_{t+j}) \ge 0 ,
\end{align}
\end{subequations}
where $\mathbf{x}_{t}$ is the state estimate provided by the UKF without the biases and with angular velocity included, $\mathbf{u}_{t}$ are individual motor thrusts, $f(\cdot)$ denotes the nominal dynamics~\cite{saviolo2022pitcn}, and the final inequality encodes the second-order CBF conditions (as derived in the collision avoidance Subsection~\ref{subsection:collision_avoidance}).

\subsection{Mode Switching} \label{subsec:switching_mode}
Mode switching is handled through a hysteresis policy on the estimator uncertainty. At each cycle $t$, the tracking UKF provides a covariance matrix $P_t$ over the target-relative state, and its trace $\operatorname{tr}(P_t)$ is used as a scalar confidence measure:
\begin{equation}
\begin{aligned}
    \textsc{loiter} \!\to\! \textsc{track} &: \quad \operatorname{tr}(P_t) \le \tau_{\uparrow}, \\
    \textsc{track} \!\to\! \textsc{loiter} &: \quad \operatorname{tr}(P_t) \ge \tau_{\downarrow},
\end{aligned}
\end{equation}
with $\tau_{\uparrow} < \tau_{\downarrow}$ enforcing hysteresis.  

While in loitering, tentative detections can still bias the heading $\psi_t^{\mathrm{loiter}}$ toward the target without triggering a full switch. Only when estimator uncertainty falls below $\tau_{\uparrow}$ does the system re-anchor $\mathcal{R}_t$ to the target and begin tracking. Conversely, if uncertainty exceeds $\tau_{\downarrow}$, the system reverts to loitering. At each transition, the UKF and NMPC are reinitialized with the new frame definition and reference trajectory, ensuring consistent state propagation and planning. Because obstacle-avoidance constraints are recomputed from depth data at every cycle, CBF constraints remain valid.


\begin{figure*}[t]
    \centering
    \includegraphics[width=\linewidth]{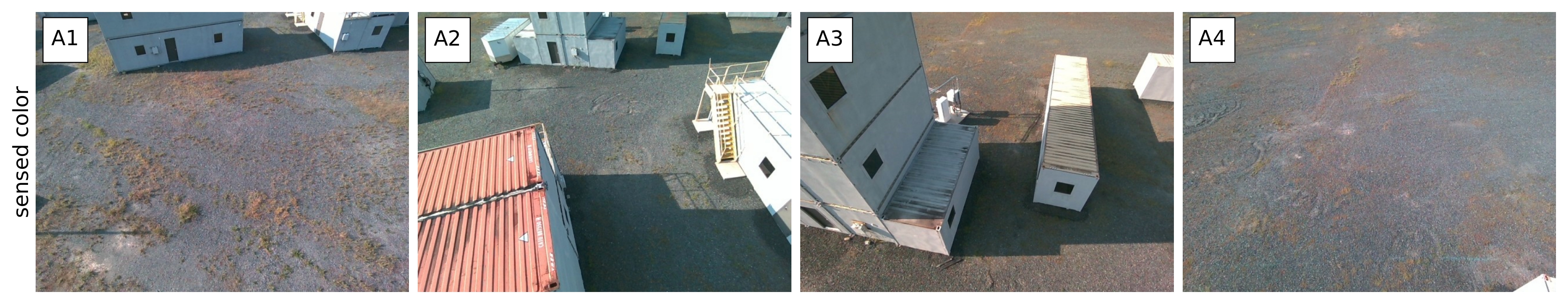}\\
    \includegraphics[width=\linewidth]{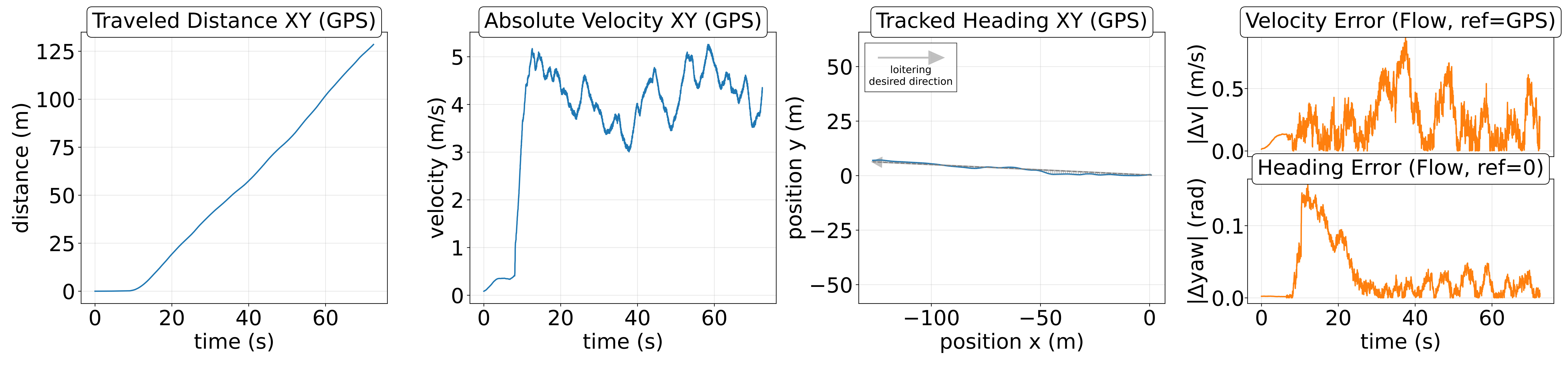}\\
    \includegraphics[width=\linewidth]{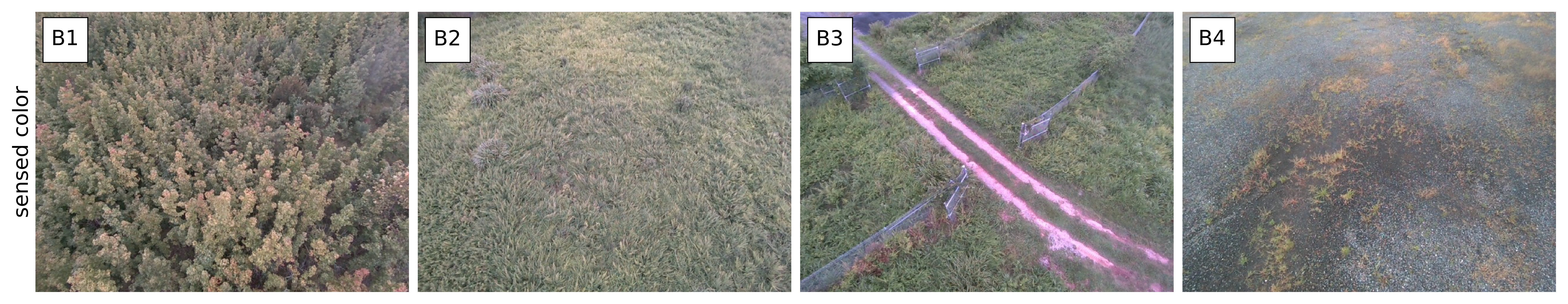}\\
    \includegraphics[width=\linewidth]{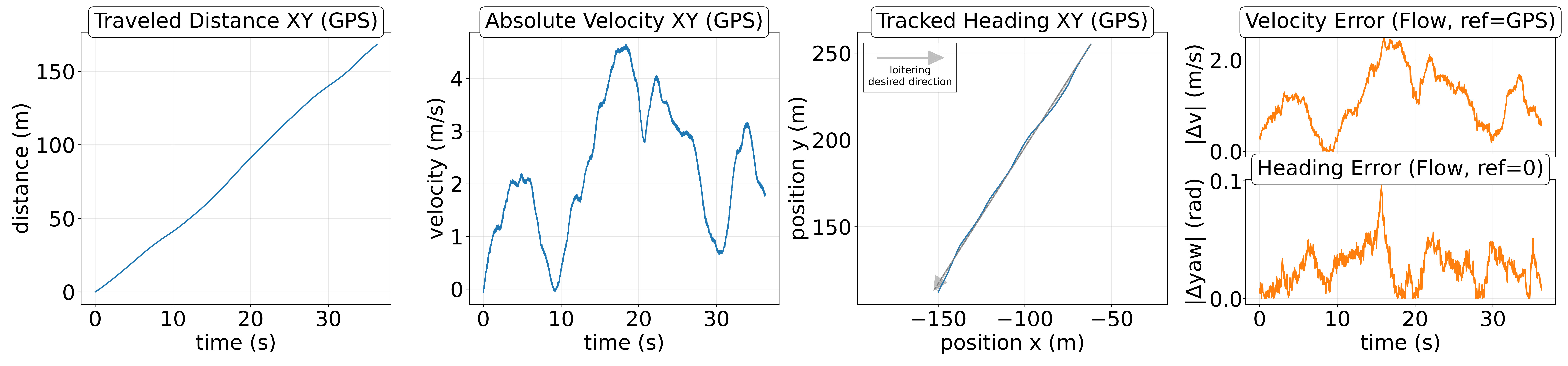}
    \caption{\textbf{Loitering experiments without a target.}
    \textit{Top}: Drift-free traversal above simulated city blocks. \textit{Bottom}: Transition flight from forest canopy into the same city environment. In both experiments, UAV maintains steady velocity and heading using only instantaneous observables, with velocity and heading errors remaining bounded relative to GPS ground truth.}
    \label{fig:loitering_above}
    \vspace{-1em}
\end{figure*}

\section{Experiments} \label{sec:exp_relnav}
We evaluate HUNT in a range of real-world environments designed to stress its ability to loiter safely, track dynamic targets, and switch seamlessly between modes. Test sites include both structured settings (urban container compounds, open fields) and unstructured natural environments (dense forest canopy, narrow forest trails). Scenarios combine static clutter with dynamic elements such as off-road vehicles and human mannequins to emulate search-and-rescue conditions.  

\subsection{Experimental Setup} \label{subsec:exp_setup_relnav}
All experiments are conducted with a quadrotor weighing $1.2~\si{kg}$, with a thrust-to-weight ratio of $4$ and a motor span of $25~\si{cm}$. Onboard computation is provided by an NVIDIA Jetson Orin NX, while low-level stabilization runs on a PixRacer Pro. The PixRacer’s IMU and barometer, centrally mounted and coincident with the body frame $\mathcal{B}$, provide attitude, angular velocity, and altitude.  
Perception uses an Intel RealSense D455 operating at $320 \times 240$ fused with monocular disparity from DepthAnythingV2~\cite{yang2024depth2} (ViT-S) accelerated on TensorRT at $19~\si{ms}$. Target detection is performed by a custom-trained YOLOv11 network, following the work~\cite{saviolo2025nova}. The camera is mounted downward at $45^{\circ}$ when flying above canopy or buildings, and forward-facing when operating at ground level to maximize the number of features in view.

The NMPC runs at $200~\si{Hz}$ with a $2~\si{s}$ prediction horizon and $N=10$ shooting steps, solved with sequential quadratic programming in \texttt{acados}~\cite{acados}. At each cycle, $K=10$ high-risk obstacle points are selected based on minimal time-to-collision~\cite{saviolo2025reactive}. CBF constraints are updated at $60~\si{Hz}$.

\begin{figure*}[t]
    \centering
    \includegraphics[width=0.495\linewidth, trim=0 0 0 0, clip]{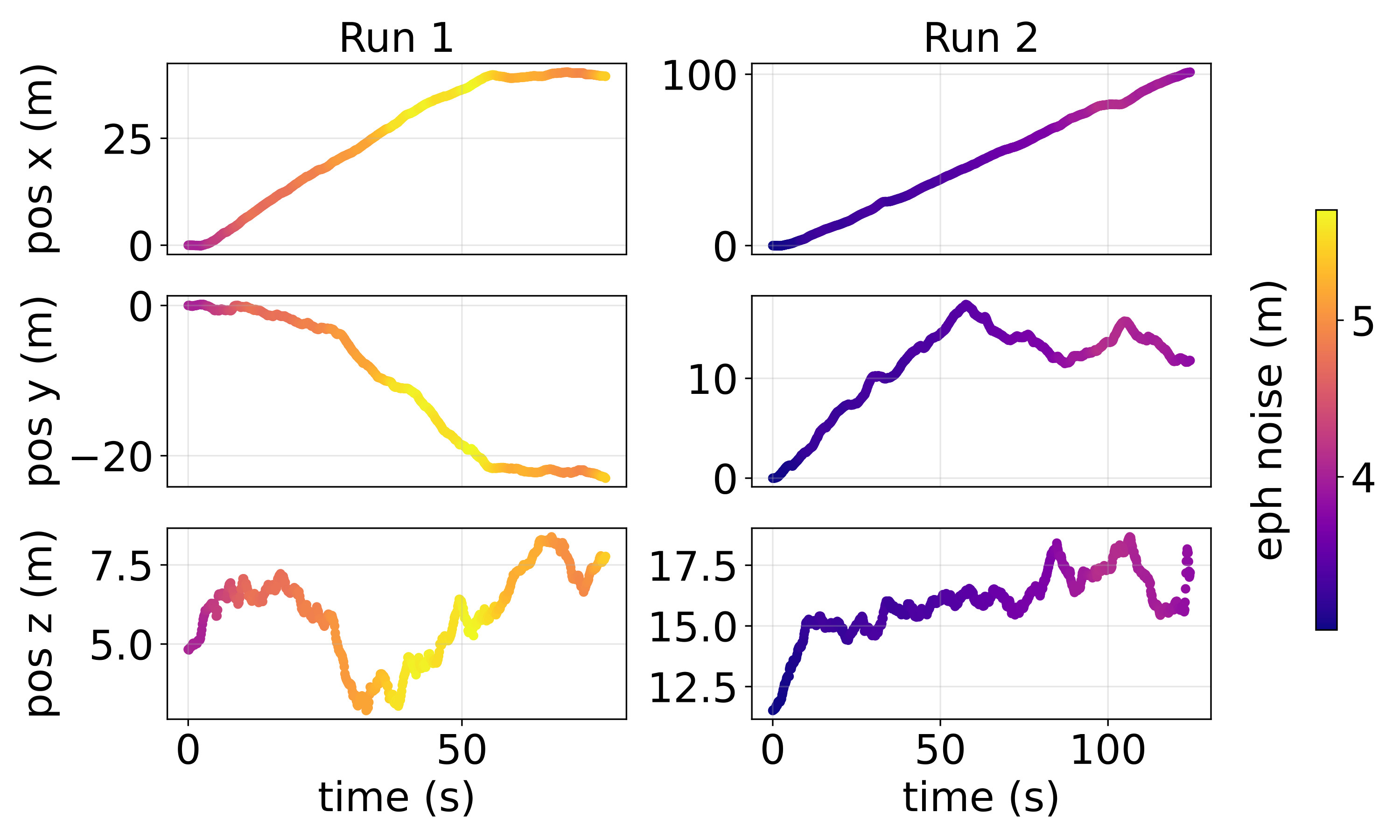}
    \includegraphics[width=0.495\linewidth, trim=160 50 40 50, clip]{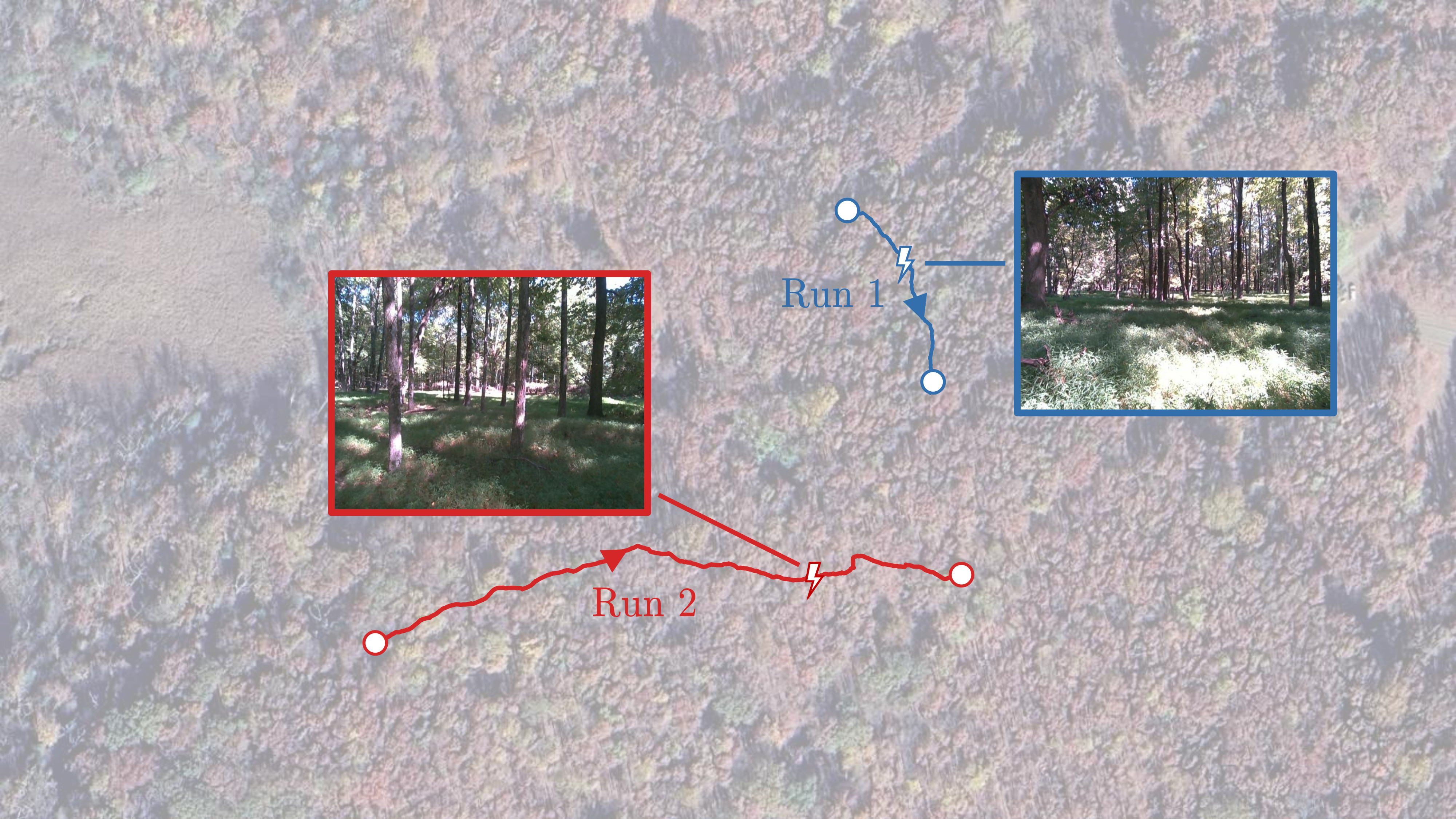}
    \caption{\textbf{Loitering under dense forest canopy.} 
    \textit{Left}: UAV trajectories for two representative runs, with GPS-reported position over time colored by estimated horizontal error (EPH). Despite errors in the order of $2$–$6~\si{m}$ that would make map-based navigation unreliable, HUNT sustains drift-free flight by relying only on instantaneous observables to guide it to maximum free space.
    \textit{Right}: satellite view of the experimental site with reconstructed flight paths. Insets illustrate the camera views.}
    \label{fig:obs_avoidance}
    \vspace{-1em}
\end{figure*}

\subsection{Loitering Above Structured and Natural Environments}
We first evaluate HUNT in the loitering regime, where no target is present and the UAV must sustain safe forward flight while continuously searching (Figure~\ref{fig:loitering_above}). At mission start, the yaw reference is initialized to the vehicle’s takeoff heading $\psi^{\mathrm{loiter}}$, the altitude reference is ramped to $p_z^{\mathrm{loiter}} = 12~\si{m}$, and forward motion begins at the commanded cruise speed $v^{\mathrm{loiter}} = 4.0~\si{m/s}$. To maximize ground coverage, the RGB-D camera is pitched downward by $45^{\circ}$.  

In the first experiment, the UAV flies above an emulated urban environment. As shown in the top row of Figure~\ref{fig:loitering_above}, the vehicle maintains steady forward progress at the commanded cruise speed, with measured velocity averaging $4.1~\si{m/s}$ and varying less than $0.3~\si{m/s}$. Heading error relative to GPS ground truth remains bounded within $2^{\circ}$.  

The second experiment repeats the mission in a more challenging mixed scenario, transitioning from flight above a dense forest canopy into structured city blocks. Despite the abrupt change in geometry and texture, the UAV preserves stable forward motion and heading traces (bottom row of Figure~\ref{fig:loitering_above}), showing that HUNT generalizes seamlessly across drastically different conditions.

\subsection{Safely Traversing Dense Unknown Forests}
We next evaluate HUNT in loitering mode below a dense forest canopy, where clutter and degraded GPS make global localization unreliable (Figure~\ref{fig:obs_avoidance}). In this setting, the RGB-D camera is mounted forward-facing. The altitude reference is set to $p_z^{\mathrm{loiter}} = 2.0~\si{m}$ and forward speed to $v^{\mathrm{loiter}} = 2.0~\si{m/s}$, while the yaw reference $\psi_t^{\mathrm{loiter}}$ is updated online at every cycle as the direction of maximum free space. Free space is computed by extracting the largest connected component in the completed depth map and selecting the point farthest from its boundary, yielding a heading that steers the UAV through the widest corridor.  

Figure~\ref{fig:obs_avoidance} presents two representative runs. The UAV sustains stable forward motion while weaving between trees, consistently maintaining clearance margins above $1.0~\si{m}$. The reactive CBF-based avoidance ensures dead ends are bypassed and collisions averted even in dense clutter. By contrast, GPS logs—recorded only for evaluation—exhibit horizontal errors of $2$–$6~\si{m}$ and spurious altitude jumps. Such errors would prevent any GPS-based approach from constructing a reliable map and would likely lead to unsafe planning and eventual crashes.



\begin{table}[t]
    \centering
    \caption{
    \textbf{Quantitative performance across SAR missions.}
    Each mission stresses a different capability: crossing a urban layout and acquiring a stationary vehicle wreckage (top row of Figure~\ref{fig:mode_switching_results}), transitioning from high-altitude search to dynamic human pursuit in a city environment (middle row), and traversing dense forest to locate and lock onto a target (bottom row). Metrics include UAV forward velocity (mean and peak), control effort (mean and peak linear acceleration), and pitch angle (mean and peak). HUNT maintains safe and agile navigation in all settings without parameter retuning.
    }
    \label{tab:mode_switching_results}
    \begin{tabular}{lcccccc}
    \toprule\toprule
    \multirow{2}{*}{\textbf{Mission}} & \multicolumn{2}{c}{\textbf{Velocity (m/s)}} & \multicolumn{2}{c}{\textbf{Effort (m/s²)}} & \multicolumn{2}{c}{\textbf{Pitch (°)}} \\
    \cmidrule(lr){2-3} \cmidrule(lr){4-5} \cmidrule(lr){6-7}
    & \textbf{Mean} & \textbf{Max} & \textbf{Mean} & \textbf{Max} & \textbf{Mean} & \textbf{Max} \\
    \midrule
    Urban Search  & 8.85 & 12.8 & 10.31 & 22.35 & 2.00 & 15.30 \\
    Urban Pursuit & 7.72 & 11.4 & 10.69 & 15.49 & 6.03 & 19.74 \\
    Forest Search & \textit{6.52}\textsuperscript{*} & \textit{9.50}\textsuperscript{*} & 10.63 & 31.11 & 1.90 & 11.31 \\
    \bottomrule\bottomrule
    \end{tabular}
    \footnotesize \textsuperscript{*}\textit{GPS velocity under canopy is unreliable due to signal degradation.}
\end{table}

\begin{figure*}[t]
    \centering
    \includegraphics[width=0.95\linewidth]{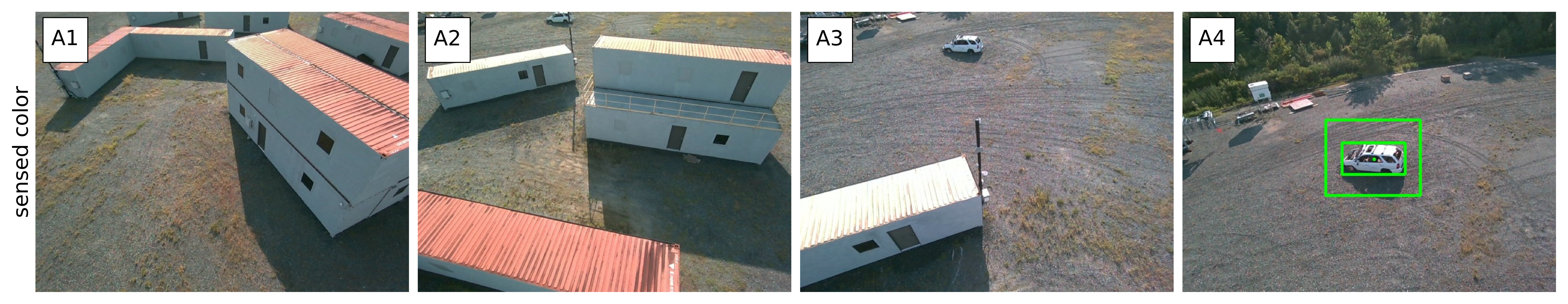}\\
    \includegraphics[width=0.95\linewidth]{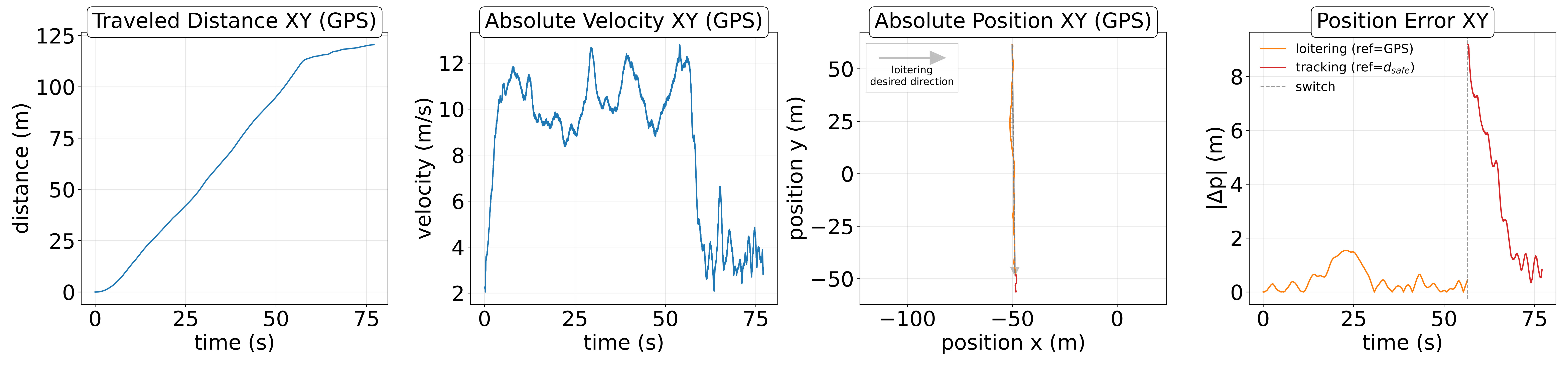}\\
    \includegraphics[width=0.95\linewidth]{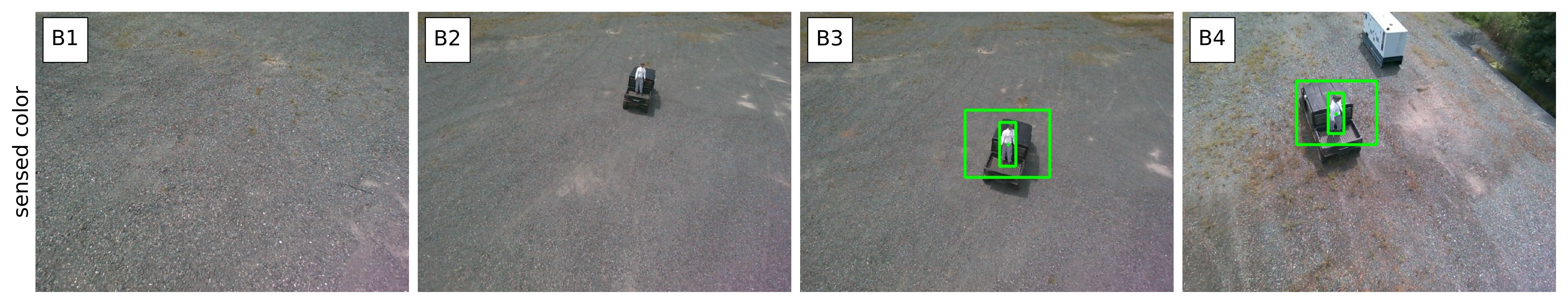}\\
    \includegraphics[width=\linewidth]{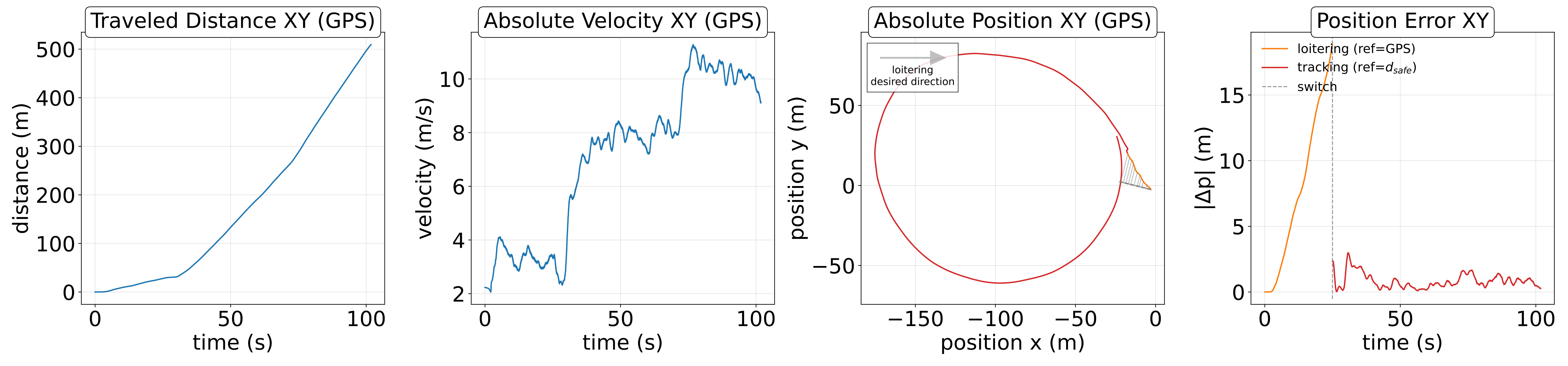}\\
    \includegraphics[width=0.95\linewidth]{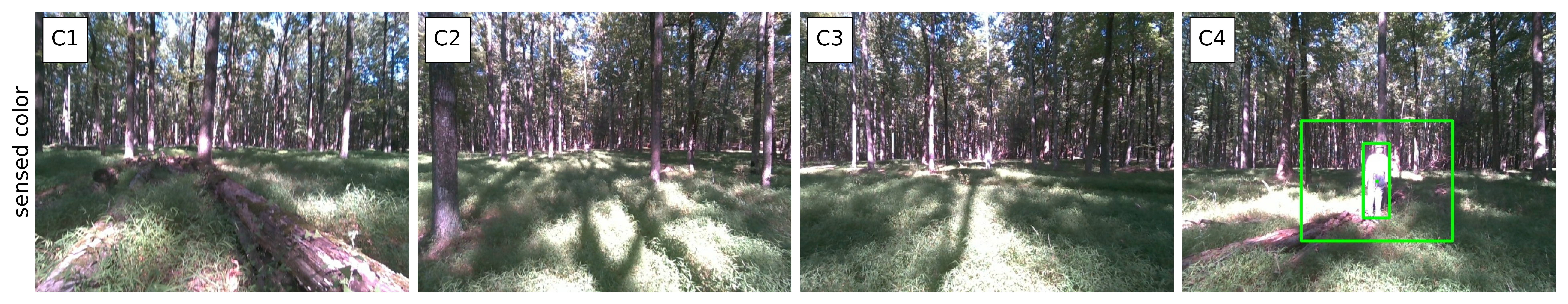}\\
    \includegraphics[width=\linewidth]{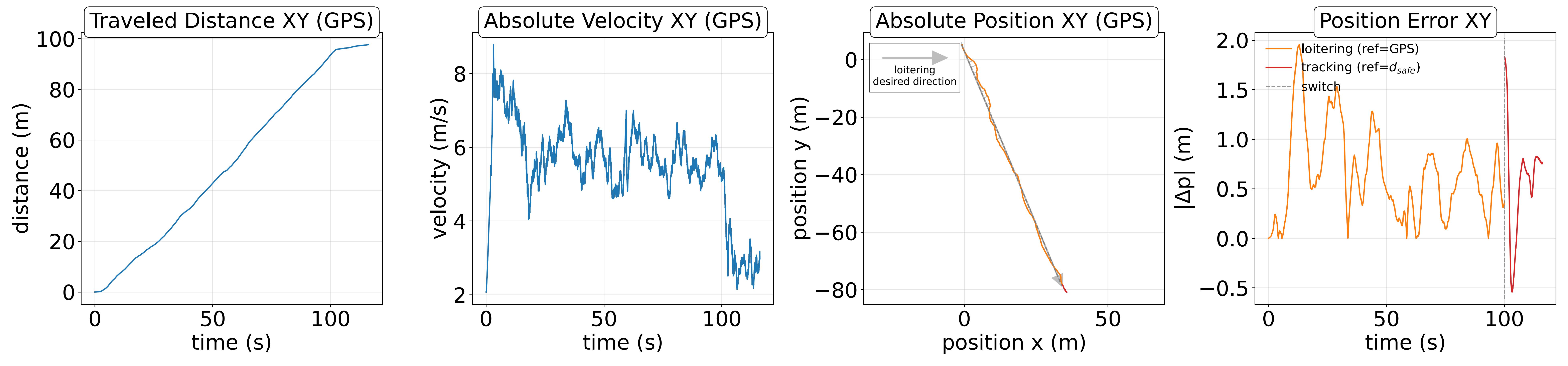}
    \caption{\textbf{Mode switching between loitering and tracking across three SAR missions.} \textit{Top (Urban Search)}: UAV flies above a simulated city, loiters while searching, then transitions into tracking upon detecting a stationary car wreckage. \textit{Middle (Urban Pursuit)}: UAV surveys a city environment, acquires a mannequin on an ATV, and maintains pursuit along irregular gravel paths with stable error bounds. \textit{Bottom (Forest Search)}: UAV loiters below dense forest canopy with active obstacle avoidance, then switches to tracking a mannequin despite clutter and occlusion.}
    \label{fig:mode_switching_results}
\end{figure*}

\subsection{Complete SAR Missions: Traversal, Acquisition, Tracking}
We finally evaluate HUNT in three full SAR missions that require seamless transitions between loitering and tracking (Figure~\ref{fig:mode_switching_results}, Table~\ref{tab:mode_switching_results}). In all cases, the UAV begins in loitering mode with fixed yaw reference $\psi^{\mathrm{loiter}}$ aligned to the initial heading at takeoff. Upon reliable detection, $\mathcal{R}_t$ is re-anchored on the target and the system switches into tracking.  

The first mission (\textit{Urban Search}) emulates locating a stationary vehicle wreckage in an urban layout. The UAV loiters with altitude $p_z^{\mathrm{loiter}} = 12~\si{m}$ and forward speed $v^{\mathrm{loiter}} = 5~\si{m/s}$ until the car is detected. After switching, the displacement error remains bounded within $\pm 0.5~\si{m}$, with plots confirming drift-free traversal before acquisition and consistent regulation thereafter.  

The second mission (\textit{Urban Pursuit}) mimics reconnaissance against a moving human target in a city. The UAV surveys with the same loitering parameters ($p_z^{\mathrm{loiter}} = 12~\si{m}$, $v^{\mathrm{loiter}} = 5~\si{m/s}$, fixed $\psi^{\mathrm{loiter}}$) until detecting a mannequin mounted on an all terrain vehicle (ATV). HUNT maintains pursuit at velocities up to $11.4~\si{m/s}$, keeping lateral error confined to a $1.2~\si{m}$ corridor despite irregular gravel paths and dynamic motion. Notably, before the confidence threshold for switching is reached, low-confidence detections bias $\psi_t^{\mathrm{loiter}}$ toward the target. This is evident in the absolute GPS traces, where the trajectory gradually bends toward the target direction during loitering and then transitions smoothly into tracking once confidence surpasses the threshold. This heading bias mechanism effectively steers the robot toward potential targets even when the initial heading is misaligned.  

The third mission (\textit{Forest Search}) represents a rescue scenario in natural clutter. The UAV loiters below canopy with altitude $p_z^{\mathrm{loiter}} = 2~\si{m}$ and forward speed $v^{\mathrm{loiter}} = 2~\si{m/s}$, following its initial heading while weaving through dense trees. Once a mannequin enters view, the system switches into tracking and sustains pursuit despite severe reflections and heading corrections up to $25^{\circ}$, while preserving safe clearance enforced by CBF constraints. The same heading bias behavior observed in the urban pursuit mission is also present in this case, though less pronounced and more clearly visible in the attached multimedia material.  

\section{Discussion And Future Works} \label{sec:discussion}
HUNT lays the foundation for instantaneous relative navigation in unstructured environments, showing that both traversal and tracking can be expressed without global pose or mapping. A central design choice is the selection of the yaw reference $\psi_t^{\mathrm{loiter}}$ during loitering, which directly influences how effectively a target can be acquired while traversing. In our experiments, we demonstrated that a fixed initial heading is sufficient for safe forward flight, since yaw drift is minimal in the absence of global orientation. We further showed that this heading can be biased by low-confidence detections, effectively steering the UAV toward potential targets, or adapted online based on free-space analysis, yielding more exploratory but also erratic behavior.  

Future work will extend these strategies with semantic reasoning from large vision–language models that can run at low frequency but provide high-level guidance. For instance, in an indoor search-and-rescue scenario after an earthquake, semantics such as recognizing windows or doors could guide loitering into unexplored regions, improving coverage without requiring explicit maps.  

Another important future direction is the introduction of lightweight short-term memory. Since HUNT operates without global pose or persistent maps, the UAV may risk loitering in cycles. A memory of recently traversed landmarks would allow the system to bias navigation away from revisited areas and toward novel regions, improving efficiency without abandoning the instantaneous relative formulation.  


\section{Conclusion} \label{sec:conclusion}
This work presented HUNT, a real-time framework for high-speed unmanned aerial navigation and tracking that eliminates reliance on global pose estimation. By formulating objectives and constraints in an instantaneous relative frame, HUNT unifies loitering, acquisition, and tracking within a single perception–control pipeline. Extensive field experiments across structured, semi-structured, and natural environments demonstrated fast traversal, reactive obstacle avoidance, and robust mode switching, achieving reliable autonomy where GPS is unreliable.  

\bibliographystyle{IEEEtran}
\bibliography{references}

\end{document}